\documentclass{article}

\usepackage{arxiv}

\usepackage[utf8]{inputenc} 
\usepackage[T1]{fontenc}    
\usepackage{hyperref}       
\usepackage{url}            
\usepackage{booktabs}       
\usepackage{amsfonts}       
\usepackage{nicefrac}       
\usepackage{microtype}      
\usepackage{lipsum}
\usepackage{graphicx}
\graphicspath{ {./images/} }

\title{Data-Driven Detection and Evaluation of Damages in Concrete Structures: Using Deep Learning and Computer Vision}

\author{
 Saeid Ataei \\
  Department of Systems and Enterprises\\
  Stevens Institute of Technology\\
  Hoboken, NJ 07030 \\
  \texttt{sataei@stevens.edu} \\
   \And
 Saeed Adibnazari \\
  Department of Aerospace Engineering\\
  Sharif University of Technology\\
  Tehran, Iran\\
  \texttt{adib@sharif.edu} \\
  \And
 Seyyed Taghi Ataei \\
  College of Interdisciplinary Science and Technology\\
  University of Tehran\\
  Tehran, Iran \\
  \texttt{st.ataei@ut.ac.ir} \\
}

\begin{document}
\maketitle
\begin{abstract}
Structural integrity is vital for maintaining the safety and longevity of concrete infrastructures such as bridges, tunnels, and walls. Traditional methods for detecting damages like cracks and spalls are labor-intensive, time-consuming, and prone to human error. To address these challenges, this study explores advanced data-driven techniques using deep learning for automated damage detection and analysis. Two state-of-the-art instance segmentation models, YOLO-v7 instance segmentation and Mask R-CNN, were evaluated using a dataset comprising 400 images, augmented to 10,995 images through geometric and color-based transformations to enhance robustness.
The models were trained and validated using a dataset split into 90\% training set, validation and test set 10\%. Performance metrics such as precision, recall, mean average precision (mAP@0.5), and frames per second (FPS) were used for evaluation. YOLO-v7 achieved a superior mAP@0.5 of 96.1\% and processed 40 FPS, outperforming Mask R-CNN, which achieved a mAP@0.5 of 92.1\% with a slower processing speed of 18 FPS.
The findings recommend YOLO-v7 instance segmentation model for real-time, high-speed structural health monitoring, while Mask R-CNN is better suited for detailed offline assessments. This study demonstrates the potential of deep learning to revolutionize infrastructure maintenance, offering a scalable and efficient solution for automated damage detection.
\end{abstract}

\keywords{Concrete Structures \and Data-Driven \and Deep Learning \and Damage Detection \and YOLO-v7 \and Mask R-CNN \and Instance Segmentation}

\section{Introduction}
\label{sec:introduction}
Structural health monitoring (SHM) is essential for ensuring the safety, efficiency, and longevity of infrastructure. Periodic inspections are particularly critical for concrete structures, such as bridges, building walls, tunnels, and dams, to identify potential defects that may compromise structural integrity. Common indicators of structural health deterioration include defects like cracks and spalls \cite{ref-gatti}. These issues often arise due to fatigue and periodic loading, underscoring the need for early, timely, and automated diagnostic methods \cite{ref-kim}. Implementing proactive measures can significantly reduce financial costs and mitigate human safety risks associated with structural failures.

Concrete is among the most widely used construction materials due to its affordability, versatility, and extensive applications in civil works and infrastructure development. Consequently, advancements in monitoring and maintaining its structural integrity are paramount. The rise of data-driven methods for analyzing and predicting structural defects represents a paradigm shift from traditional manual inspections \cite{ref-jahanshahi1}. These methodologies, grounded in verifiable data analysis rather than intuition or subjective judgment, provide a robust framework for deriving insights and developing solutions \cite{ref-jahanshahi2}. Such methods enable predictive assessments by leveraging knowledge from historical and current data, thereby reducing biases and false assumptions.

Recent advancements in machine learning and artificial intelligence, particularly deep learning, have transformed SHM processes. Vision-based deep learning algorithms can efficiently identify structural defects, including cracks and spalls, with superior accuracy compared to manual inspections. These automated methods expedite the diagnosis process, decrease reliance on specialized expertise, and prevent potential human errors \cite{ref-badrinarayanan, ref-long, ref-cha}. By automating the detection process, deep learning reduces time and costs while enhancing the reliability of defect identification \cite{ref-dung}.

\subsection{Problem Definition}
Data-driven techniques for defect detection in concrete structures are gaining traction due to advancements in hardware and software technologies. These methods can be categorized into two groups:

\begin{enumerate}
    \item \textbf{Non-intelligent Monitoring:} Traditional approaches, such as visual inspections and manual experiments, heavily depend on human expertise, leading to safety risks, inefficiencies, and inaccuracies. These methods are often labor-intensive and time-consuming \cite{ref-bigdeli}.
    
    \item \textbf{Intelligent Monitoring:} Advanced methodologies, including image processing, machine learning, and deep learning, address the limitations of traditional approaches. These techniques facilitate cost-effective, precise, and automated monitoring of structural health \cite{ref-heidari}.
\end{enumerate}

Concrete SHM often involves managing large datasets. Techniques like instance segmentation in deep learning help in extracting meaningful patterns from this data. By employing vision-based algorithms, structural damage can be accurately identified, classified, and localized, ensuring timely interventions. Moreover, the ability to utilize pre-trained models and transfer learning has significantly enhanced the accessibility and efficiency of deep learning-based SHM systems \cite{ref-azimi}.

\subsection{Methods for Detection}
Deep learning, a subset of machine learning, encompasses various learning paradigms, including supervised, semi-supervised, and unsupervised learning \cite{ref-xu}. These approaches require collecting, preprocessing, and preparing data before applying algorithms for classification and pattern extraction. Techniques such as transfer learning, augmented datasets, and pre-trained networks (e.g., the SDNE2018 dataset, which contains 56,000 images of bridges and walls) expedite the training process and improve accuracy \cite{ref-silva}.

Classification and segmentation methods are crucial in SHM. Classification involves categorizing data into distinct classes, such as cracks, spalls, or intact regions. Instance segmentation extends classification by identifying individual defects within a broader context, providing detailed insights into structural health \cite{ref-cha2}. Additionally, few-shot learning methods address data limitations by focusing on learning patterns from minimal data samples. This innovative approach has proven effective across diverse fields, including image classification, natural language processing, and structural damage detection \cite{ref-majdi}.

\subsection{State of the Research Field}
Concrete defect detection has evolved from heuristic and machine vision techniques to sophisticated deep learning methods. Traditional image processing filters, such as Canny and Sobel, were once prevalent but limited in handling complex structural defects \cite{ref-yeum}. The advent of artificial neural networks (ANNs) brought significant advancements, with convolutional neural networks (CNNs) leading the charge \cite{ref-badrinarayanan}.

CNN-based techniques bypass manual feature extraction, enabling direct learning of image features. For instance, Cha and Choi \cite{ref-cha2} demonstrated the efficacy of deep CNN models under variable brightness and shadow conditions. Similarly, studies employing hybrid models—merging CNNs with traditional methods like Otsu filtering—achieved higher accuracy in detecting cracks and predicting their direction \cite{ref-majdi}. Recent research also explores transformer-based models, which, despite higher computational costs, exhibit enhanced flexibility and reduced bias compared to CNNs \cite{ref-li}.

Several pioneering works have focused on classification and segmentation. Noushin Bigdeli et al. \cite{ref-bigdeli} utilized a CNN-based approach to classify concrete cracks into distinct categories, achieving an impressive 99.3\% accuracy. Other studies employed advanced pre-trained networks, such as ResNet and AlexNet, for transfer learning, significantly improving defect detection precision \cite{ref-azimi, ref-xu2}. For instance, the ResNet-50 model—renowned for its depth and accuracy—has been instrumental in concrete damage classification \cite{ref-he}.

Object detection algorithms like YOLO and R-CNN are gaining momentum for their ability to localize and categorize structural defects. Zhou et al. \cite{ref-xu2} proposed a faster R-CNN model to identify and locate seismic damages in reinforced concrete columns with an 80\% accuracy. Similarly, Maedeh et al. \cite{ref-maeda} used single-stage object detection networks, such as SSD-MobileNet, to identify road damages with a 75\% accuracy.

\subsection{Diverging Hypotheses and Controversies}
While deep learning techniques promise unprecedented accuracy, there are debates surrounding their scalability and computational efficiency. Transformer-based models, despite their high potential, pose challenges due to elevated computational demands. Moreover, data-driven approaches often rely on the availability of large, high-quality datasets. Inconsistent data labeling and variability in imaging conditions further complicate model generalization \cite{ref-li}.

Another point of contention is the effectiveness of semantic versus instance segmentation methods. While semantic segmentation identifies damage categories, instance segmentation distinguishes between multiple defects within the same category. Researchers like Mousavi and Bakhshi \cite{ref-mousavi} advocate for encoder-decoder architectures like ResNet101 coupled with U-Net for enhanced segmentation, achieving 99.39\% accuracy.

\subsection{Significance and Purpose of Study}
This study aims to advance SHM through innovative deep learning methodologies, leveraging combined datasets to enhance generalizability. By focusing on instance segmentation, the research prioritizes accurate localization and classification of damages. The proposed methods offer:

\begin{itemize}
    \item Improved accuracy in identifying structural defects.
    \item Automated detection with minimal human intervention.
    \item Enhanced safety through early defect identification and resolution.
    \item Cost-effectiveness by minimizing reliance on manual inspections \cite{ref-mousavi}.
\end{itemize}

The exploration of deep learning—spanning CNNs, object detection, and instance segmentation—has yielded substantial advancements in SHM. By integrating existing datasets and applying instance segmentation, this study highlights the potential for achieving unprecedented precision in defect detection. Moreover, it demonstrates how combining novel algorithms and big data analytics can significantly improve the resilience and reliability of critical infrastructure \cite{ref-guo}.

\section{Research Methodology}
\label{sec:methodology}

This section delineates the detailed methodology employed in this study, aimed at detecting cracks and spalls in concrete structures using advanced deep learning models. The steps cover dataset preparation, data augmentation, model training, and evaluation procedures. The approaches were developed to ensure replicability and consistency for researchers wishing to replicate or extend this study's findings.

\subsection{Dataset Preparation}

The dataset for this study was curated from three publicly available sources to enhance generalizability and robustness. Specifically:

\begin{itemize}
    \item 100 images from the Özgenel segmentation dataset \cite{ref-ozgenel} with a resolution of 3024×4032.
    \item 100 images from a referenced dataset \cite{ref-shawn} with resolutions ranging from 296×306 to 334×306.
    \item 200 images of cracks and spalls from another dataset \cite{ref-zhang} with resolutions of 768×768 and 960×1280.
\end{itemize}

To enhance the generalizability of the network results, we combined a selection of images from three datasets referenced in \cite{ref-ozgenel,ref-zhang,ref-shawn}. These images can be downloaded from the \href{https://www.kaggle.com/datasets/stmlen/cconcrack}{CConCrack dataset on Kaggle} and are available for researchers to use. (If you use this dataset, please cite this research paper.) The combined dataset totaled 400 images, expanded through augmentation to 10,995 images. All images were resized to a uniform 640×640 resolution for preprocessing. Table \ref{tab:data_distribution} provides the data distribution after augmentation:

\begin{table}
  \caption{Dataset Split}
  \centering
  \begin{tabular}{ccc}
    \toprule
    \textbf{Data}       & \textbf{Training} & \textbf{Validation, Test} \\
    \midrule
    \textbf{Number}     & 9000              & 1000                         \\
    \bottomrule
  \end{tabular}
  \label{tab:data_distribution}
\end{table}

\begin{figure}
    \centering
    \includegraphics[width=1\linewidth]{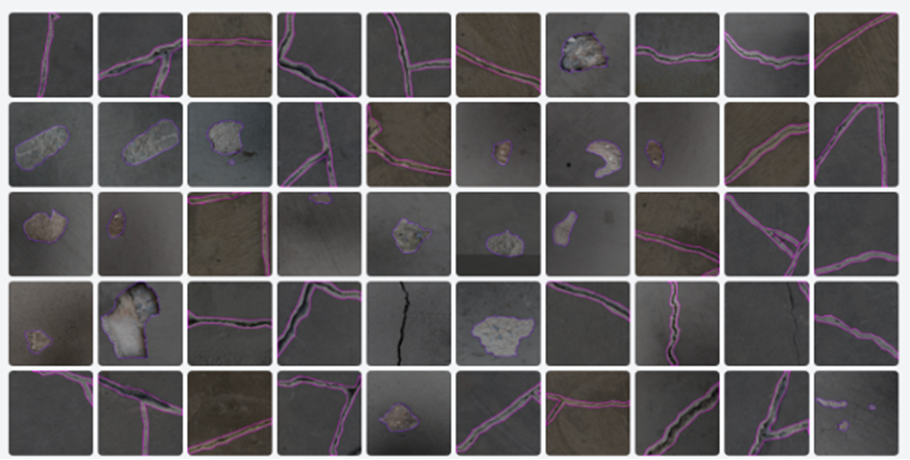}
    \caption{A Sample of Labeled Data}
    \label{fig:labeled_data}
\end{figure}

Following data collection, Roboflow was used for labeling tasks at the instance level, ensuring accuracy in data categorization. Example labeled data is illustrated in Figure \ref{fig:labeled_data}.

\subsection{Data Augmentation}

Deep learning models often require substantial data to avoid overfitting. Data augmentation techniques were employed in two stages:

1. Initial Augmentation using the RoboFlow tool:
   \begin{itemize}
       \item Flip: Horizontal and vertical.
       \item 90° Rotation: Clockwise and counterclockwise.
       \item Cutting: 0\% minimum zoom, 20\% maximum zoom.
   \end{itemize}

   Table \ref{tab:roboflow_methods} shows the methods applied using RoboFlow.

   \begin{table}
     \caption{RoboFlow Data Augmentation Methods}
     \centering
     \begin{tabular}{ll}
       \toprule
       \textbf{Method}      & \textbf{Operation}                \\
       \midrule
       Flip                  & Horizontal, Vertical             \\
       Rotation              & ±90°                             \\
       Cutting               & Min zoom 0\%, Max zoom 20\%     \\
       \bottomrule
     \end{tabular}
     \label{tab:roboflow_methods}
   \end{table}

2. Model-Specific Augmentation for YOLO-v7:
   \begin{itemize}
       \item HSV Adjustments: Hue, saturation, and brightness.
       \item Image Displacement: Translation along X and Y axes.
       \item Dimension Changes: Zooming in/out.
       \item Mosaic: Combining four images.
   \end{itemize}

   Table \ref{tab:yolo7_methods} shows augmentation settings for YOLO-v7.

   \begin{table}
     \caption{YOLO-v7 Data Augmentation Methods}
     \centering
     \begin{tabular}{lll}
       \toprule
       \textbf{Method}           & \textbf{Effect}                      & \textbf{Value}               \\
       \midrule
       HSV                       & Hue, saturation, brightness         & H: 0.015, S: 0.7, V: 0.4  \\
       Displacement              & Translation (X or Y axis)          & 0.2                         \\
       Dimension Change          & Zoom in/out                        & 0.5                         \\
       Rotate                    & ± Directional Turn                 & 0.5                         \\
       Mosaic                    & Combine Four Images                & 1                           \\
       \bottomrule
     \end{tabular}
     \label{tab:yolo7_methods}
   \end{table}

\subsection{Mask R-CNN Training}

The Mask R-CNN model was utilized to detect and segment cracks and spalls. The steps for training included:

\begin{enumerate}
    \item Installing Detectron2, a state-of-the-art platform developed by Facebook AI.
    \item Downloading and mapping the dataset.
    \item Defining training parameters, such as a basic learning rate of 0.00025 and a batch size of 2 images per category.
    \item Training for a maximum of 100,000 iterations with an early stop at 40,000 if no cost function improvement was observed.
    \item Testing the trained model on unseen images and saving the model’s weights.
\end{enumerate}

The loss function included classification, bounding box regression, and binary cross-entropy components. Transfer learning was employed using weights pre-trained on the COCO dataset, followed by fine-tuning to adapt to the new dataset.

\subsection{YOLO-v7 Instance-Segmentation Training}

The YOLO-v7 instance segmentation model, another high-performance object detection model, was trained in this study using the following methodology:

\begin{enumerate}
    \item Setting up a GPU-enabled environment.
    \item Installing model requirements.
    \item Preparing and formatting the dataset for YOLO-v7.
    \item Training with an initial learning rate of 0.01, optimized by stochastic gradient descent.
    \item Evaluating model performance on a validation set.
    \item Testing on unseen images using transferred weights pre-trained on the COCO dataset.
\end{enumerate}

Parameters for YOLO-v7 training are summarized in Table \ref{tab:yolo7_parameters}:

\begin{table}
    \caption{YOLO-v7 Training Parameters}
    \centering
    \begin{tabular}{ll}
        \toprule
        \textbf{Parameter}            & \textbf{Value}             \\
        \midrule
        Repetitions                  & 200                         \\
        Batch Size                   & 16                          \\
        Classes                      & 2 (cracks, spalls)         \\
        Initial Learning Rate        & 0.01                        \\
        End Learning Rate            & 0.1                         \\
        Momentum                     & 0.937                       \\
        Weight Optimizer             & 0.0005                      \\
        IoU Threshold                & 0.2                         \\
        \bottomrule
    \end{tabular}
    \label{tab:yolo7_parameters}
\end{table}

\subsection{Validation and Testing}

Validation and testing play crucial roles in assessing model robustness. The dataset was divided into three categories as shown in Table \ref{tab:data_distribution}. Metrics utilized include:

\begin{itemize}
    \item Mean Average Precision (mAP): To assess precision across multiple categories.
    \item Intersection over Union (IoU): For evaluating segmentation accuracy.
\end{itemize}

Both the Mask R-CNN and YOLO-v7 instance segmentation models exhibited high performance on unseen test data, indicating robust generalizability. Examples of predictions and comparisons with ground truth images are illustrated in the Results section.

By utilizing advanced deep learning architectures—Mask R-CNN and YOLO-v7 instance segmentation—alongside effective dataset preparation and augmentation techniques, the research demonstrates a scalable and reproducible approach to improving the accuracy of structural health monitoring systems. Detailed validation ensures that findings are robust, and transfer learning maximizes efficiency while reducing computational costs. Future researchers are encouraged to build upon this methodology, exploring additional neural architectures and expansive datasets for even greater precision in SHM applications.

\section{Results}
\label{sec:results}

This section presents the experimental results obtained from training and evaluating two deep learning models, Mask R-CNN and YOLO-v7 instance segmentation, for detecting cracks and spalls in concrete structures. The findings are divided into subheadings, providing an in-depth interpretation and conclusions based on the outcomes.

\subsection{Evaluation Metrics}

Evaluating the effectiveness of object detection models involves several critical metrics:

\begin{itemize}
    \item \textbf{Precision}: The proportion of correctly identified positive samples relative to the total positive samples predicted by the model.
    \item \textbf{Recall}: The fraction of actual positives identified correctly.
    \item \textbf{Intersection over Union (IoU)}: A metric to measure the overlap between the predicted and ground truth bounding boxes, with values beyond a 0.5 threshold considered true positives.
    \item \textbf{Mean Average Precision (mAP)}: The average precision across multiple IoU thresholds (e.g., 0.5).
    \item \textbf{Frames Per Second (FPS)}: An evaluation of the model’s inference speed, critical for real-time applications.
\end{itemize}

Table \ref{tab:confusion_matrix} illustrates the confusion matrix components utilized in these assessments.

\begin{table}
    \caption{Confusion Matrix Components}
    \centering
    \begin{tabular}{ccc}
        \toprule
        \textbf{Real Label} & \textbf{Predicted Label} & \textbf{Description} \\ 
        \midrule
        True (+)            & True (+)                 & True Positive (TP)   \\ 
        False (-)          & True (+)                 & False Positive (FP)  \\ 
        True (+)            & False (-)                & False Negative (FN)  \\ 
        False (-)          & False (-)                & True Negative (TN)   \\ 
        \bottomrule
    \end{tabular}
    \label{tab:confusion_matrix}
\end{table}

\subsection{Mask R-CNN Results}

The Mask R-CNN model was trained with 100,000 iterations and achieved notable accuracy metrics. The results are summarized in Table \ref{tab:mask_rcnn_accuracy} and Table \ref{tab:class_specific_mask_rcnn} below.

1. Accuracy Metrics:

\begin{table}
    \caption{Training Accuracy Metrics for Mask R-CNN}
    \centering
    \begin{tabular}{ccc}
        \toprule
        \textbf{Metric}      & \textbf{AP50} & \textbf{AP75} \\ 
        \midrule
        Bounding Box         & 95.5\%        & 86.3\%        \\ 
        Classification       & 92.1\%        & 55.8\%        \\ 
        \bottomrule
    \end{tabular}
    \label{tab:mask_rcnn_accuracy}
\end{table}

2. Recall Metrics: \\ 
   AR@IoU 0.5: 62.8\%

3. Class-Specific Metrics:

\begin{table}
    \caption{Class-Specific Metrics for Mask R-CNN}
    \centering
    \begin{tabular}{cc}
        \toprule
        \textbf{Class} & \textbf{AP} \\ 
        \midrule
        Crack          & 37.3\%      \\ 
        Spall          & 79.3\%      \\ 
        \bottomrule
    \end{tabular}
    \label{tab:class_specific_mask_rcnn}
\end{table}

4. Inference Speed: The detection speed was recorded at 18 FPS (55 milliseconds), which is suboptimal for real-time detection.

\begin{figure}
    \centering
    \includegraphics[width=0.5\linewidth]{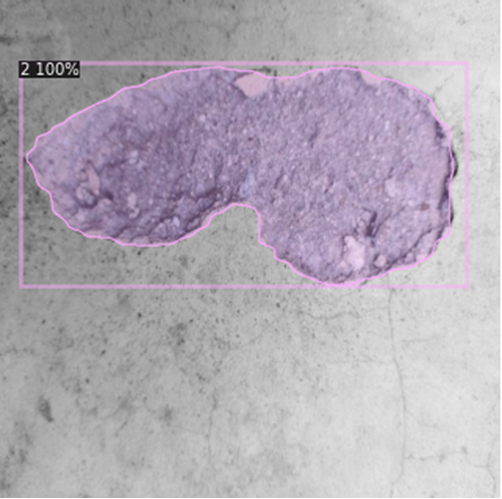}
    \caption{Sample Mask R-CNN Output for Spall}
    \label{fig:mask_rcnn_output}
\end{figure}

In Fig. \ref{fig:mask_rcnn_output}, a sample Mask R-CNN output for the Spall class is shown.

\subsection{YOLO-v7 Instance Segmentation Results}

YOLO-v7 instance segmentation displayed exceptional performance and faster inference times during evaluation, as shown in Table \ref{tab:yolo7_accuracy} below.

1. \textbf{Accuracy Metrics:}
   \begin{itemize}
    \item mAP@0.5: 96.1\%
    \item Crack Detection: mAP@0.5 = 92.7\%
    \item Spall Detection: mAP@0.5 = 99.5\%
\end{itemize}

\begin{table}
    \caption{Accuracy Metrics for YOLO-v7 Instance Segmentation}
    \centering
    \begin{tabular}{cccc}
        \toprule
        \textbf{Metric}      & \textbf{Precision} & \textbf{Recall} & \textbf{mAP50} \\ 
        \midrule
        Overall              & 94.9\%            & 94.3\%         & 96.1\%        \\ 
        Crack                & 94.5\%            & 88.6\%         & 92.7\%        \\ 
        Spall                & 95.2\%            & 99.9\%         & 99.5\%        \\ 
        \bottomrule
    \end{tabular}
    \label{tab:yolo7_accuracy}
\end{table}

2. \textbf{Inference Speed:}
\begin{itemize}
    \item Speed: 40 FPS (24.9 milliseconds), suitable for real-time applications.
\end{itemize}

\begin{figure}
    \centering
    \includegraphics[width=0.5\linewidth]{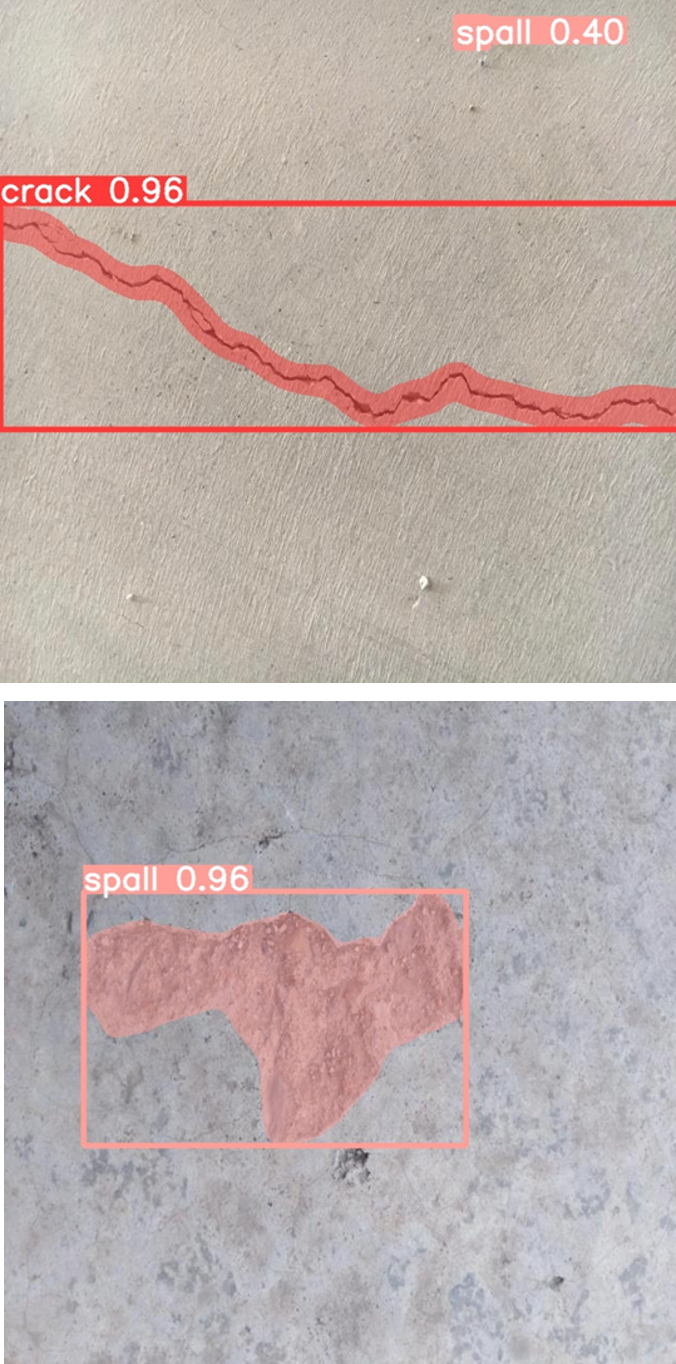}
    \caption{YOLO-v7 Instance Segmentation Model Predictions}
    \label{fig:yolo7_predictions}
\end{figure}

\begin{figure}
    \centering
    \includegraphics[width=0.5\linewidth]{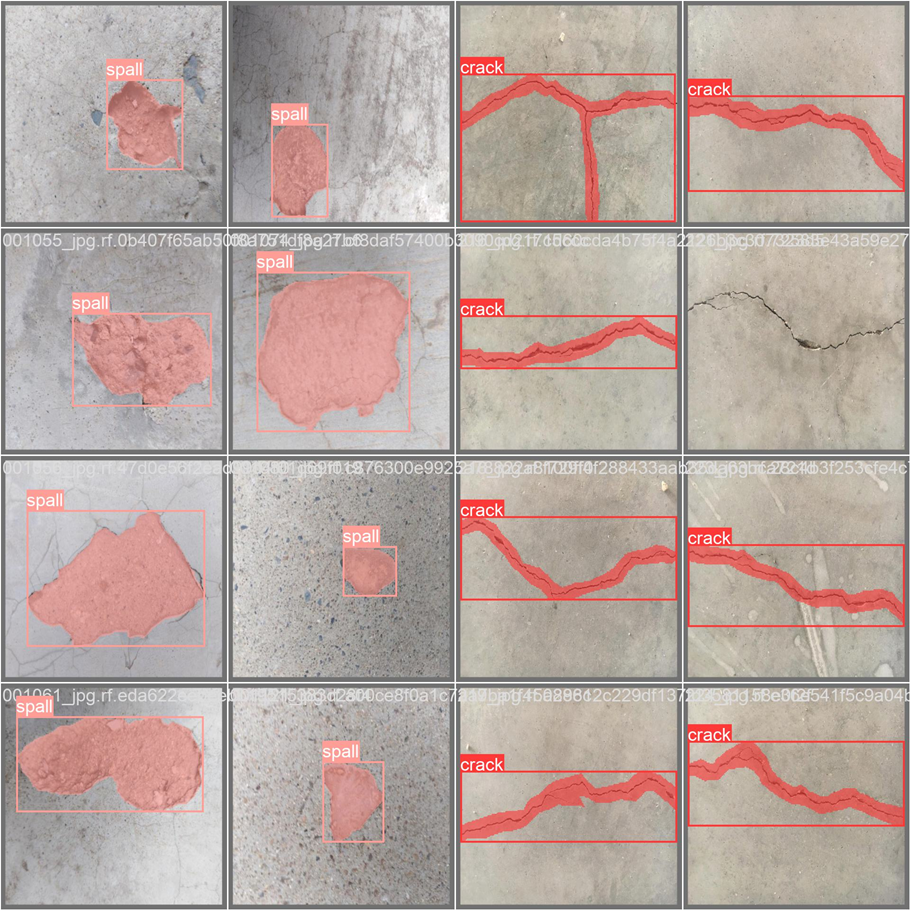}
        \includegraphics[width=0.5\linewidth]{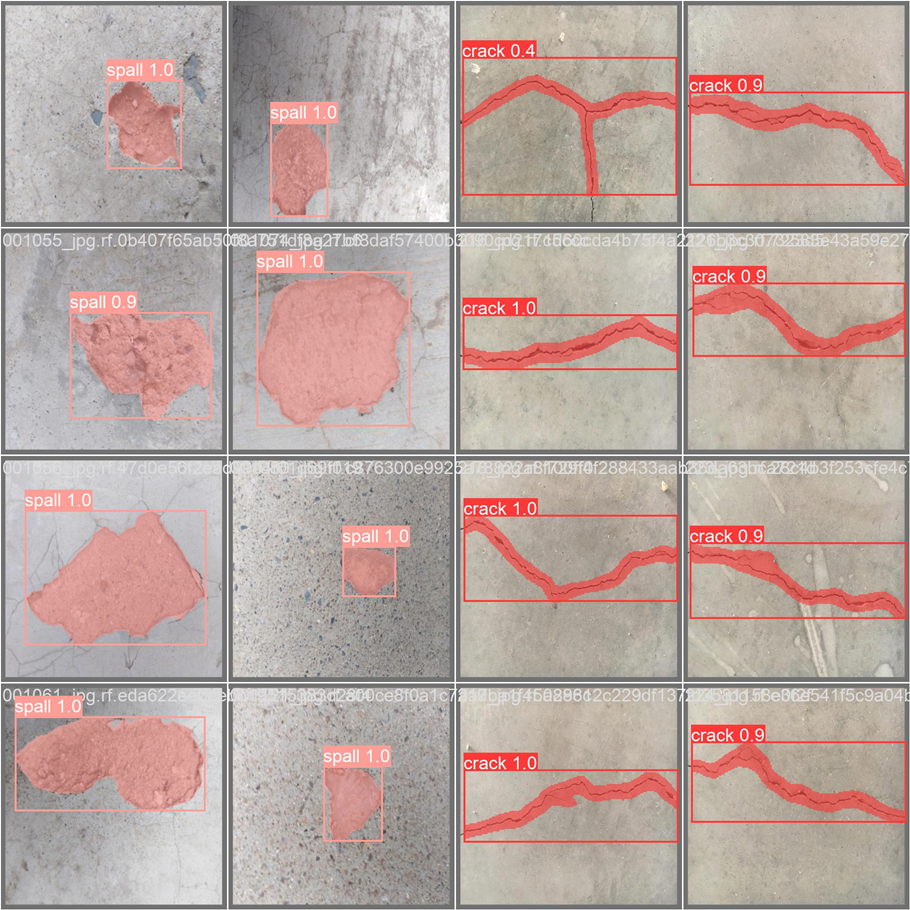}
    \caption{Ground Truth (above) vs. Predicted Labels (below)}
    \label{fig:ground_truth_vs_predicted}
\end{figure}

In Fig. \ref{fig:yolo7_predictions}, a YOLO-v7 instance segmentation model prediction is shown, and in Fig. \ref{fig:ground_truth_vs_predicted}, the ground truth vs. predicted labels can be seen.

3. \textbf{Model Iterations:}
\begin{itemize}
    \item Optimal accuracy was achieved at 168 iterations.
\end{itemize}

\begin{figure}
    \centering
    \includegraphics[width=0.5\linewidth]{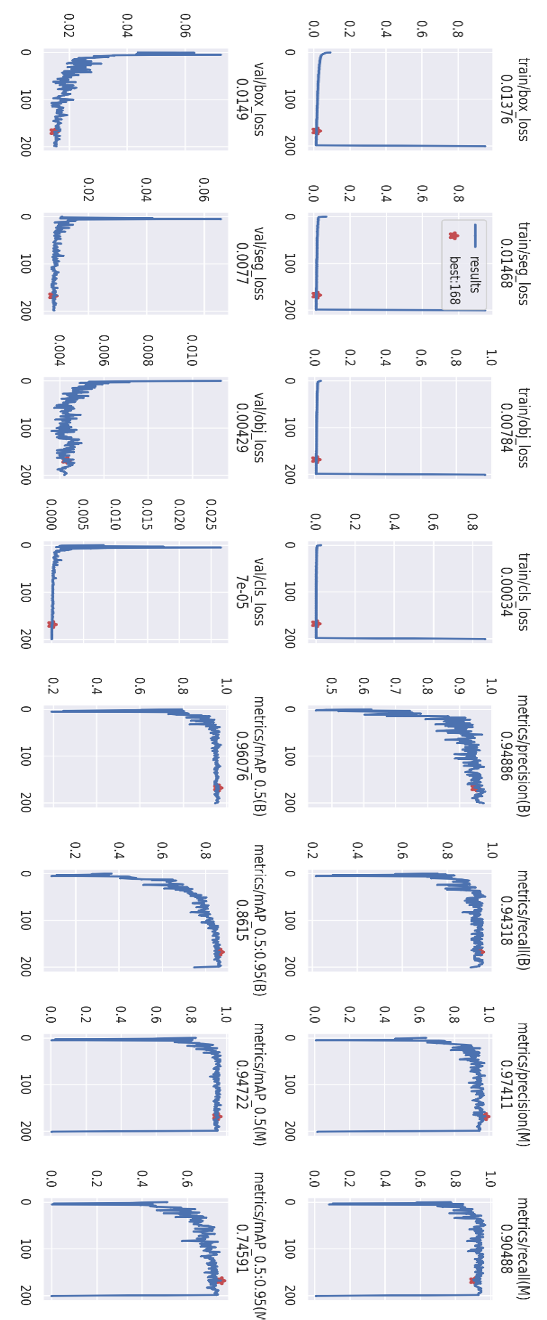}
    \caption{YOLO-v7 Instance Segmentation Model Training Performance}
    \label{fig:yolo7_training_performance}
\end{figure}

Figure \ref{fig:yolo7_training_performance} shows YOLO-v7 instance segmentation model training performance.

\subsection{Comparative Analysis}

1. \textbf{Performance Metrics:}
\begin{itemize}
    \item While both models excelled in identifying spall defects, the YOLO-v7 instance segmentation model exhibited higher overall mAP@0.5 scores and faster inference times, as shown in Table \ref{tab:comparative_metrics}, making it more suitable for real-time applications.
\end{itemize}

\begin{table}
    \caption{Comparative Metrics for Mask R-CNN and YOLO-v7 Instance Segmentation Models}
    \centering
    \begin{tabular}{ccc}
        \toprule
        \textbf{Model}       & \textbf{mAP50} & \textbf{FPS} \\ 
        \midrule
        Mask R-CNN           & 92.1\%         & 18          \\ 
        YOLO-v7 Instance Segmentation & 96.1\%         & 40          \\ 
        \bottomrule
    \end{tabular}
    \label{tab:comparative_metrics}
\end{table}

2. \textbf{Model Strengths:}
\begin{itemize}
    \item Mask R-CNN: Superior precision for crack classification.
    \item YOLO-v7 Instance Segmentation: Overall higher mAP and superior speed for practical deployment.
\end{itemize}

\begin{figure}
    \centering
    \includegraphics[width=0.9\linewidth]{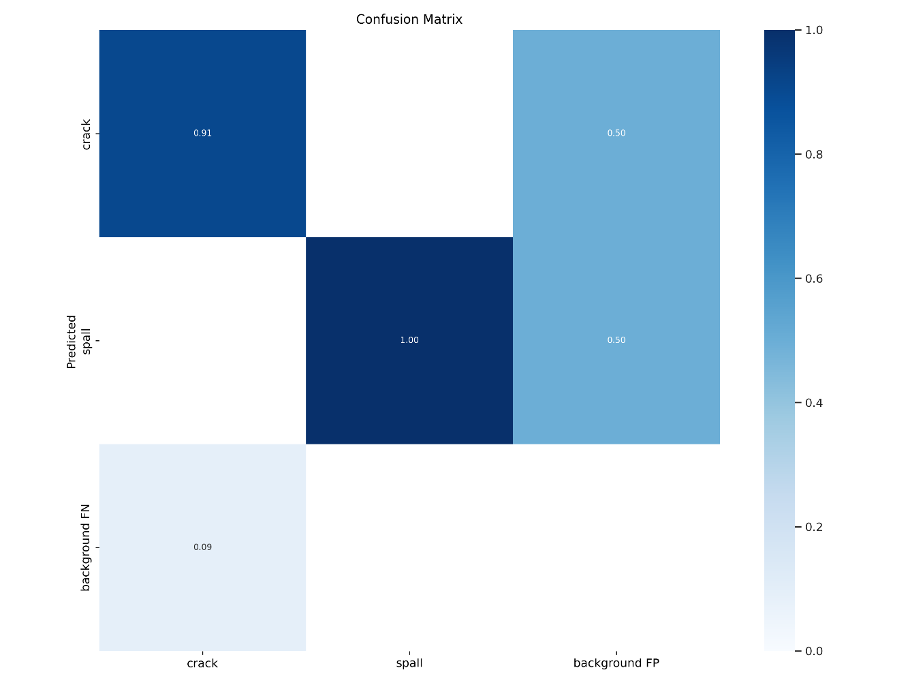}
    \caption{YOLO-v7 Instance Segmentation Confusion Matrix}
    \label{fig:yolo7_confusion_matrix}
\end{figure}

3. \textbf{Confusion Matrix Analysis:}
\begin{itemize}
    \item YOLO-v7 Instance Segmentation achieved a 91\% accuracy for crack detection and 100\% for spall detection (Fig. \ref{fig:yolo7_confusion_matrix}).
\end{itemize}

The findings highlight the effectiveness of the YOLO-v7 instance segmentation model over Mask R-CNN in terms of both accuracy and speed. YOLO-v7’s real-time capability and exceptional mAP metrics underscore its practicality for structural health monitoring tasks. These results pave the way for deploying the YOLO-v7 instance segmentation in critical infrastructure monitoring, ensuring timely defect detection and reducing inspection costs.

\subsection{Discussion}
This study trained and evaluated two instance segmentation models, Mask R-CNN and YOLO-v7 instance segmentation, focusing on their applicability for structural damage detection. The comparative analysis based on metrics such as mAP50, precision, recall, speed, and testing on real-world images and videos revealed distinct advantages and limitations of each approach.

\subsection{Performance Metrics and Accuracy}
The results demonstrate that YOLO-v7 instance segmentation outperformed Mask R-CNN in terms of accuracy metrics. As summarized in Table \ref{tab:model_comparison} below, the mAP50, precision, and recall for YOLO-v7 instance segmentation were 96.1, 94.9, and 94.3, respectively, compared to Mask R-CNN’s scores of 92.1, 92.0, and 62.8.

\begin{table}
    \centering
    \caption{Comparison of Two Instance Segmentation Models}
    \begin{tabular}{cccc}
        \toprule
        \textbf{Model} & \textbf{mAP50} & \textbf{Precision} & \textbf{Recall}\\
        \midrule
        Mask R-CNN & 92.1 & 92.0 & 62.8\\
        YOLO-v7 & 96.1 & 94.9 & 94.3\\
        \bottomrule
    \end{tabular}
    \label{tab:model_comparison}
\end{table}

This data is also visualized in Figure \ref{fig:accuracy_comparison} below:

\begin{figure}
    \centering
    \includegraphics[width=0.8\linewidth]{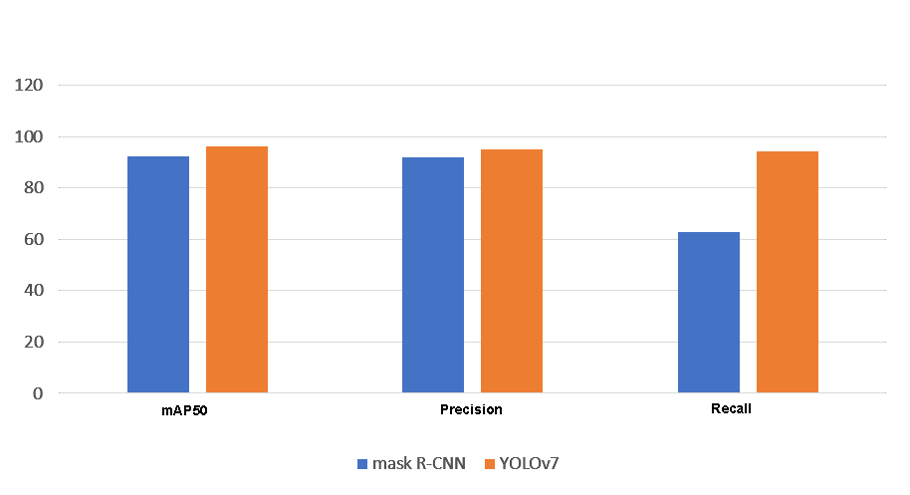}
    \caption{Comparison of the Accuracy of the Two Models}
    \label{fig:accuracy_comparison}
\end{figure}

These findings align with prior research, which has shown that YOLO models tend to excel in tasks requiring high recall and precision due to their end-to-end architecture and optimized loss functions designed for real-time performance. This makes YOLO-v7 instance segmentation particularly suitable for detecting subtle damages in concrete structures where recall is critical to minimize undetected defects.

\subsection{Speed and Real-Time Performance}
The speed comparison, as illustrated in Table \ref{tab:speed_comparison} and Figure \ref{fig:speed_comparison} below, underscores YOLO-v7 instance segmentation’s superiority for real-time applications. With an average runtime of 24.9 milliseconds and a frame rate of 40 fps, YOLO-v7 instance segmentation is nearly twice as fast as Mask R-CNN, which has a runtime of 55 milliseconds and achieves only 18 fps.

\begin{table}
    \caption{Comparison of the Speed of the Two Instance Segmentation Models}
    \centering
    \begin{tabular}{ccc}
        \toprule
        \textbf{Model}       & \textbf{Time (msec)} & \textbf{Speed (fps)} \\ 
        \midrule
        Mask R-CNN           & 55                   & 18                   \\ 
        YOLO-v7 Instance Segmentation & 24.9                 & 40                   \\ 
        \bottomrule
    \end{tabular}
    \label{tab:speed_comparison}
\end{table}

\begin{figure}
    \centering
    \includegraphics[width=0.8\linewidth]{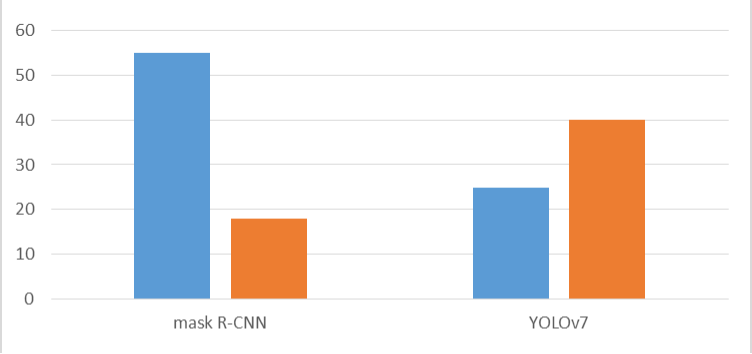}
    \caption{Comparing Speed (orange frames/sec) and Runtime (milliseconds)}
    \label{fig:speed_comparison}
\end{figure}

The faster processing speed of YOLO-v7 instance segmentation is attributable to its streamlined network architecture and computational efficiency. This efficiency is crucial for applications in structural health monitoring, where real-time detection can significantly reduce response times and enhance safety during critical events, such as earthquakes or structural collapses.

\subsection{Generalization to Unseen Data}
To test generalization, the models were evaluated using random images and videos sourced from the internet, which were not part of the training dataset. Figures \ref{fig:random_photo} and \ref{fig:random_video} below illustrate the detection capabilities of YOLO-v7 instance segmentation, showcasing its robustness in identifying damages in novel scenarios.

\begin{figure}
    \centering
    \includegraphics[width=0.8\linewidth]{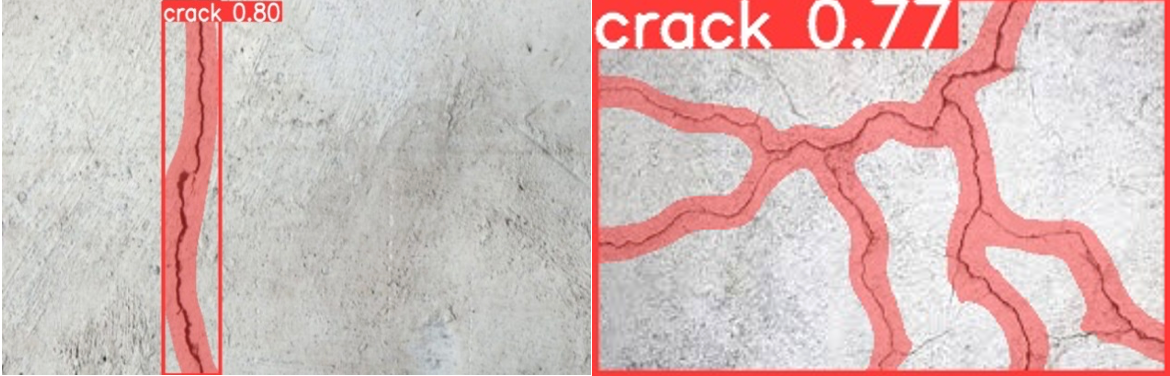}
    \caption{Testing the Accuracy of the Model with a Random Photo Taken from the Internet}
    \label{fig:random_photo}
\end{figure}

\begin{figure}
    \centering
    \includegraphics[width=1\linewidth]{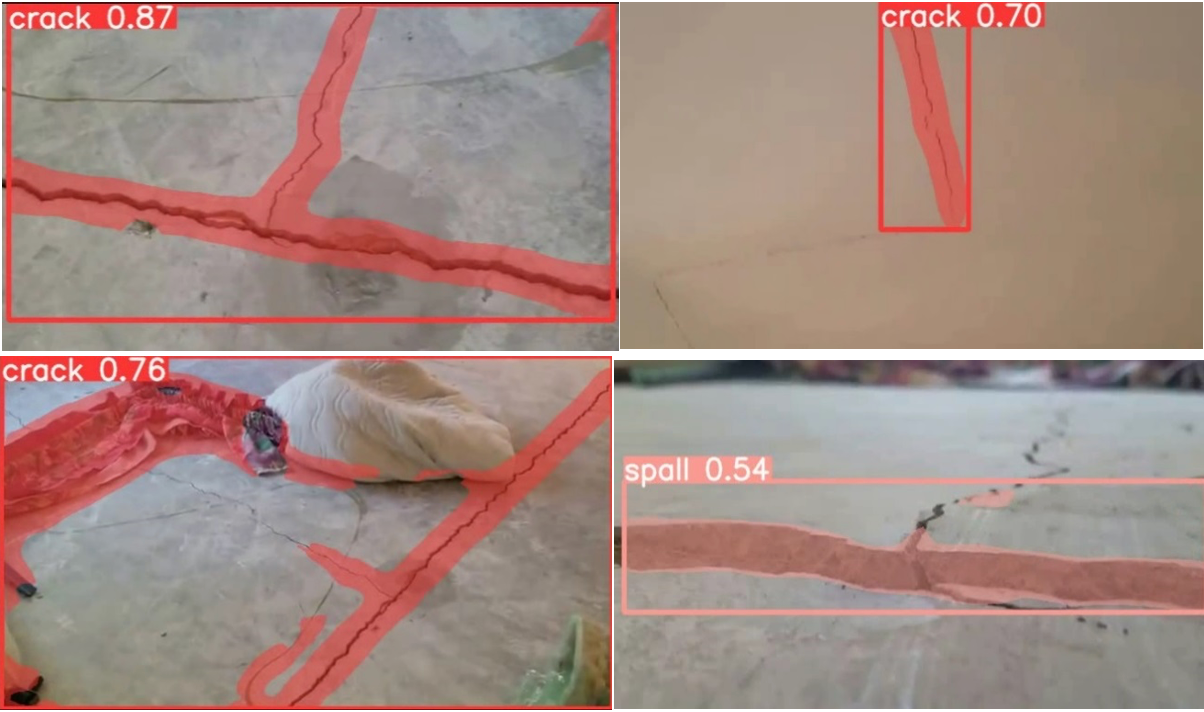}
    \caption{Testing the Accuracy of the Model with a Random Video Taken from the Internet}
    \label{fig:random_video}
\end{figure}

This robustness is essential for practical deployment in dynamic environments where pre-trained datasets may not encompass all possible variations in structural damage.

\subsection{Comparison with Other Methods}
Although a direct comparison with other methods in structural health monitoring is constrained by differences in datasets, the findings suggest that YOLO-v7 instance segmentation's approach is well-suited for damage detection tasks. Previous studies have primarily relied on classification algorithms, which, while effective for coarse damage detection, lack the granularity and localization capabilities of instance segmentation models. YOLO-v7 instance segmentation’s ability to combine high accuracy with precise localization positions it as a superior alternative for this domain.

\subsection{Broader Context and Implications}
The results of this study have broad implications for the field of structural health monitoring. The deployment of instance segmentation models like YOLO-v7 can enhance the automation and accuracy of damage detection processes, reducing reliance on manual inspections. This is particularly beneficial for large-scale infrastructure projects, where timely identification of damages can prevent catastrophic failures and optimize maintenance schedules.

The superior performance of YOLO-v7 also opens avenues for integrating AI-driven monitoring systems with Internet of Things (IoT) devices. For instance, integrating YOLO-v7 with drone-based inspection systems can facilitate real-time damage assessment of inaccessible structures, such as high-rise buildings or underwater foundations. Moreover, the model’s efficiency makes it feasible for deployment on edge devices, enabling on-site analysis without the need for high-bandwidth data transmission to centralized servers.

\subsection{Limitations and Future Research}
Despite its advantages, YOLO-v7 instance segmentation has limitations that warrant further investigation. One challenge is its reliance on annotated datasets for training. The creation of comprehensive and diverse datasets that encompass various types of structural damages remains a labor-intensive task. Additionally, while YOLO-v7 performs well in real-time scenarios, its performance in detecting extremely fine or subtle cracks may be constrained by its resolution limits.

Future research should focus on the following areas:
\begin{enumerate}
    \item \textbf{Dataset Expansion:} 
    Developing larger and more diverse datasets with detailed annotations at the instance level, including a wide range of structural damages across different materials and environmental conditions.

    \item \textbf{Enhanced Structural Health Monitoring Frameworks:} 
    By integrating the trained YOLO-v7 model into a comprehensive structural health monitoring system, tasks ranging from data collection to defect identification can be fully automated through the deployment of drones and IoT-enabled cameras.

    \item \textbf{Detection of Additional Damage Classes:} 
    Extending the model’s capabilities to identify more types of structural damages beyond cracking and collapse, such as rebar corrosion, spalling, and delamination, will enhance its utility for infrastructure management.

    \item \textbf{Model Optimization:} 
    Exploring hybrid architectures that combine YOLO-v7’s speed with Mask R-CNN’s ability to detect fine details, potentially leveraging transfer learning techniques.

    \item \textbf{Real-Time Edge Deployment:} 
    Investigating the deployment of YOLO-v7 on edge devices can optimize the performance and accessibility of real-time monitoring in field conditions.

    \item \textbf{Integration with IoT:} 
    Exploring the integration of instance segmentation models with IoT frameworks for continuous, real-time monitoring of critical infrastructure.

    \item \textbf{Explainability and Trust:} 
    Developing interpretability tools to explain model predictions can increase trust and adoption in safety-critical applications. Clear visualizations of detected defects and their severity can aid decision-making by engineers and maintenance personnel.
\end{enumerate}

To conclude, this study underscores the potential of the YOLO-v7 instance segmentation model as a transformative tool for structural health monitoring. Its high accuracy, speed, and robustness make it a promising candidate for automating damage detection and enhancing the safety and longevity of infrastructure. Addressing the identified limitations and pursuing the proposed future research directions will further harness the capabilities of AI-driven solutions to revolutionize structural health monitoring.

\section{Conclusions}

This study successfully addressed the identification of structural defects, particularly cracks and spalls, in concrete structures using two advanced instance segmentation models: Mask R-CNN and YOLO-v7. The research demonstrated that automated defect detection through these deep learning models not only enhances accuracy but also achieves real-time performance, providing a significant improvement over traditional methods, which are often labor-intensive, expensive, and potentially hazardous.

The YOLO-v7 instance segmentation model proved to be highly effective, achieving an accuracy of 96.1\% (mAP50) and a real-time detection speed of 40 frames per second. This performance represents a substantial improvement over the Mask R-CNN model, which scored lower in both accuracy and speed. The results confirm the advantages of the YOLO-v7 instance segmentation model for structural health monitoring tasks due to its speed, accuracy, and robustness.

Moreover, the YOLO-v7 model demonstrated robustness in generalization tests using unseen images and videos, confirming its potential for practical applications where the diversity of real-world data demands robust and adaptable algorithms. In comparison to previous approaches, the use of instance segmentation models marks a significant shift towards higher granularity and localization capabilities, enabling the precise identification of structural defects. The integration of such models in infrastructure monitoring systems offers a cost-effective, scalable, and efficient alternative to traditional inspection methods.
The success of the YOLO-v7 instance segmentation model in this study underscores the transformative potential of deep learning models in structural health monitoring. By providing a scalable, real-time, and highly accurate solution, such systems can reduce maintenance costs, prevent catastrophic failures, and extend the lifespan of infrastructure. The integration of these models with emerging technologies, such as IoT and drones, holds significant promise for the future of automated infrastructure management.

\bibliographystyle{unsrt}  


\begin{thebibliography}{1}

\bibitem{ref-ozgenel}
C.~F. Özgenel. Concrete crack segmentation dataset. {\em Mendeley Data} {\bf 2019}.

\bibitem{ref-zhang}
C. Zhang, C. Chang, and M. Jamshidi. Simultaneous pixel-level concrete defect detection and grouping using a fully convolutional model. {\em Structural Health Monitoring} {\bf 2021}, {\em 20}, 2199--2215.

\bibitem{ref-shawn}
Y. Shawn. FCN for crack recognition. Available online: \url{https://github.com/OnionDoctor/FCN_for_crack_recognition} (accessed on 13 March 2018).

\bibitem{ref-gatti}
M. Gatti. Structural health monitoring of an operational bridge: A case study. {\em Eng. Struct.} {\bf 2019}, {\em 195}, 200--209.

\bibitem{ref-kim}
H. Kim, E. Ahn, M. Shin, and S.-H. Sim. Crack and noncrack classification from concrete surface images using machine learning. {\em Struct. Health Monit.} {\bf 2019}, {\em 18}, 725--738.

\bibitem{ref-dorafshan1}
S. Dorafshan, M. Maguire, N.~V. Hoffer, and C. Coopmans. Fatigue crack detection using unmanned aerial systems in under-bridge inspection. {\em Ida. Transp. Dep.} {\bf 2017}, {\em 2}, 1--120.

\bibitem{ref-dorafshan2}
S. Dorafshan, R.~J. Thomas, and M. Maguire. Fatigue crack detection using unmanned aerial systems in fracture critical inspection of steel bridges. {\em J. Bridge Eng.} {\bf 2018}, {\em 23}, 04018078.

\bibitem{ref-jahanshahi1}
M.~R. Jahanshahi, J.~S. Kelly, S.~F. Masri, and G.~S. Sukhatme. A survey and evaluation of promising approaches for automatic vision-based defect detection of bridge structures. {\em Struct. Infrastruct. Eng.} {\bf 2009}, {\em 5}, 455--486.

\bibitem{ref-jahanshahi2}
M.~R. Jahanshahi and S.~F. Masri. A new methodology for non-contact accurate crack width measurement through photogrammetry for automated structural safety evaluation. {\em Smart Mater. Struct.} {\bf 2013}, {\em 22}, 035019.

\bibitem{ref-yeum}
C.~M. Yeum and S.~J. Dyke. Vision-based automated crack detection for bridge inspection. {\em Comput.-Aided Civ. Infrastruct. Eng.} {\bf 2015}, {\em 30}, 759--770.

\bibitem{ref-badrinarayanan}
V. Badrinarayanan, A. Kendall, and R. Cipolla. Segnet: A deep convolutional encoder-decoder architecture for image segmentation. {\em IEEE Trans. Pattern Anal. Mach. Intell.} {\bf 2017}, {\em 39}, 2481--2495.

\bibitem{ref-long}
J. Long, E. Shelhamer, and T. Darrell. Fully convolutional networks for semantic segmentation. In {\em Proceedings of the IEEE Conference on Computer Vision and Pattern Recognition}, Boston, MA, USA, 7--12 June 2015; pp. 3431--3440.

\bibitem{ref-cha}
Y.~J. Cha, W. Choi, G. Suh, S. Mahmoudkhani, and O. Büyüköztürk. Autonomous structural visual inspection using region-based deep learning for detecting multiple damage types. {\em Comput.-Aided Civ. Infrastruct. Eng.} {\bf 2018}, {\em 33}, 731--747.

\bibitem{ref-dung}
C.~V. Dung and L.~D. Anh. Autonomous concrete crack detection using deep fully convolutional neural network. {\em Autom. Constr.} {\bf 2019}, {\em 99}, 52--58.

\bibitem{ref-bigdeli}
N. Bigdeli, H. Jabbari, and M. Shojaei. An intelligent method for classification of cracks in concrete structures based on deep neural networks. {\em Amirkabir Journal of Civil Engineering} {\bf 2020}, {\em 53}, No. 8.

\bibitem{ref-heidari}
A. Heidari, M.~R. Pourtabari, and M.~R. Khalilianpour. Evaluation of cracking in concrete dams by artificial neural networks. In {\em 2nd International Conference on New Research Findings in Civil Engineering, Architecture and Urban Management}, Tehran, 2016.

\bibitem{ref-azimi}
M. Azimi, A.~D. Eslamlou, and G. Pekcan. Data-driven structural health monitoring and damage detection through deep learning: State-of-the-art review. {\em Sensors} {\bf 2020}, {\em 20}(10), 2778.

\bibitem{ref-xu}
Y. Xu, Y. Bao, Y. Zhang, and H. Li. Attribute-based structural damage identification by few-shot meta learning with inter-class knowledge transfer. {\em Structural Health Monitoring} {\bf 2020}.

\bibitem{ref-silva}
W.~R.~L.~d. Silva and D.~S.~d. Lucena. Concrete cracks detection based on deep learning image classification. In {\em Proceedings} {\bf 2018}, {\em 2}, 489.

\bibitem{ref-cha2}
Y.~J. Cha and W. Choi. Deep learning-based crack damage detection using convolutional neural networks. {\em Comput.-Aided Civ. Infrastruct. Eng.} {\bf 2017}, {\em 32}(5), 361--378.

\bibitem{ref-majdi}
A.~R. Majdi Flah, M.~L. Suleiman, and M.~L. Nehdi. Classification and quantification of cracks in concrete structures using deep learning vision-based techniques. {\em Cement and Concrete Composites} {\bf 2020}, {\em 114}, 103781.

\bibitem{ref-ahmed}
B. Ahmed, S. Mangalathu, and J.~S. Jeon. Seismic damage state predictions of reinforced concrete structures using stacked long short-term memory neural networks. {\em Journal of Building Engineering} {\bf 2022}, {\em 46}, 103737.

\bibitem{ref-harirchian1}
E. Harirchian, T. Lahmer, and S. Rasulzade. Earthquake hazard safety assessment of existing buildings using optimized multi-layer perceptron neural network. {\em Energies} {\bf 2020}, {\em 13}(8), 2060.

\bibitem{ref-harirchian2}
E. Harirchian, K. Jadhav, V. Kumari, and T. Lahmer. ML-EHSAPP: a prototype for machine learning-based earthquake hazard safety assessment of structures by using a smartphone app. {\em Eur. J. Environ. Civ. Eng.} {\bf 2021}, 1--21.

\bibitem{ref-he}
K. He, X. Zhang, S. Ren, and J. Sun. Deep residual learning for image recognition. In {\em Proceedings of the IEEE conference on computer vision and pattern recognition}, Las Vegas, NV, USA, 26--30 June 2016; pp. 770--778.

\bibitem{ref-xu2}
Y. Xu, S. Wei, Y. Bao, and H. Li. Automatic seismic damage identification of reinforced concrete columns from images by a region-based deep convolutional neural network. {\em Structural Control and Health Monitoring} {\bf 2019}, e2313.

\bibitem{ref-maeda}
H. Maeda, Y. Sekimoto, T. Seto, T. Kashiyama, and H. Omata. Road damage detection and classification using deep neural networks with smartphone images. {\em Computer-Aided Civil and Infrastructure Engineering} {\bf 2018}, {\em 33}(12), 1127--1141.

\bibitem{ref-li}
G. Li, B. Ma, S. He, X. Ren, and Q. Liu. Automatic tunnel crack detection based on U-Net and a convolutional neural network with alternately updated clique. {\em Sensors} {\bf 2020}, {\em 20}, 717.

\bibitem{ref-guo}
M.-H. Guo et al. Segnext: Rethinking convolutional attention design for semantic segmentation. {\em arXiv preprint} arXiv:2209.08575 (2022).

\bibitem{ref-mousavi}
M. Mousavi and A. Bakhshi. Crack detection in concrete members using encoder-decoder models based on deep learning. {\em Journal of Civil Engineering} {\bf 2022}, {\em 38.2}(2.2), 79--88.

\end{thebibliography}

\end{document}